\PassOptionsToPackage{unicode}{hyperref}
\PassOptionsToPackage{hyphens}{url}
\PassOptionsToPackage{dvipsnames,svgnames,x11names}{xcolor}
\documentclass[
  10pt,
  letterpaper,
]{article}
\usepackage{xcolor}
\usepackage[margin=0.9in]{geometry}
\usepackage{amsmath,amssymb}
\usepackage{iftex}
\ifPDFTeX
  \usepackage[T1]{fontenc}
  \usepackage[utf8]{inputenc}
  \usepackage{textcomp} 
\else 
  \usepackage{unicode-math} 
  \defaultfontfeatures{Scale=MatchLowercase}
  \defaultfontfeatures[\rmfamily]{Ligatures=TeX,Scale=1}
\fi
\usepackage{lmodern}
\ifPDFTeX\else
\fi
\IfFileExists{upquote.sty}{\usepackage{upquote}}{}
\IfFileExists{microtype.sty}{
  \usepackage[]{microtype}
  \UseMicrotypeSet[protrusion]{basicmath} 
}{}
\makeatletter
\@ifundefined{KOMAClassName}{
  \IfFileExists{parskip.sty}{%
    \usepackage{parskip}
  }{
    \setlength{\parindent}{0pt}
    \setlength{\parskip}{6pt plus 2pt minus 1pt}}
}{
  \KOMAoptions{parskip=half}}
\makeatother
\makeatletter
\ifx\paragraph\undefined\else
  \let\oldparagraph\paragraph
  \renewcommand{\paragraph}{
    \@ifstar
      \xxxParagraphStar
      \xxxParagraphNoStar
  }
  \newcommand{\xxxParagraphStar}[1]{\oldparagraph*{#1}\mbox{}}
  \newcommand{\xxxParagraphNoStar}[1]{\oldparagraph{#1}\mbox{}}
\fi
\ifx\subparagraph\undefined\else
  \let\oldsubparagraph\subparagraph
  \renewcommand{\subparagraph}{
    \@ifstar
      \xxxSubParagraphStar
      \xxxSubParagraphNoStar
  }
  \newcommand{\xxxSubParagraphStar}[1]{\oldsubparagraph*{#1}\mbox{}}
  \newcommand{\xxxSubParagraphNoStar}[1]{\oldsubparagraph{#1}\mbox{}}
\fi
\makeatother

\usepackage{longtable,booktabs,array}
\usepackage{calc} 
\usepackage{etoolbox}
\makeatletter
\patchcmd\longtable{\par}{\if@noskipsec\mbox{}\fi\par}{}{}
\makeatother
\IfFileExists{footnotehyper.sty}{\usepackage{footnotehyper}}{\usepackage{footnote}}
\makesavenoteenv{longtable}
\usepackage{graphicx}
\makeatletter
\newsavebox\pandoc@box
\newcommand*\pandocbounded[1]{
  \sbox\pandoc@box{#1}%
  \Gscale@div\@tempa{\textheight}{\dimexpr\ht\pandoc@box+\dp\pandoc@box\relax}%
  \Gscale@div\@tempb{\linewidth}{\wd\pandoc@box}%
  \ifdim\@tempb\p@<\@tempa\p@\let\@tempa\@tempb\fi
  \ifdim\@tempa\p@<\p@\scalebox{\@tempa}{\usebox\pandoc@box}%
  \else\usebox{\pandoc@box}%
  \fi%
}
\def\fps@figure{htbp}
\makeatother

\NewDocumentCommand\citeproctext{}{}

\makeatletter
 \let\@cite@ofmt\@firstofone
 \def\@biblabel#1{}
 \def\@cite#1#2{{#1\if@tempswa , #2\fi}}
\makeatother
\newlength{\cslhangindent}
\newlength{\csllabelwidth}
\newenvironment{CSLReferences}[2] 
 {\begin{list}{}{%
  \setlength{\itemindent}{0pt}
  \setlength{\leftmargin}{0pt}
  \setlength{\parsep}{0pt}
  \ifodd #1
   \setlength{\leftmargin}{\cslhangindent}
   \setlength{\itemindent}{-1\cslhangindent}
  \fi
  \setlength{\itemsep}{#2\baselineskip}}}
 {\end{list}}
\usepackage{calc}

\providecommand{\tightlist}{%
  \setlength{\itemsep}{0pt}\setlength{\parskip}{0pt}}

\ifdefined\DeclareUnicodeCharacter
  \DeclareUnicodeCharacter{2212}{\textminus}
\fi
\usepackage{array}
\usepackage{booktabs}
\usepackage{colortbl}
\usepackage{float}
\usepackage{microtype}
\usepackage{parskip}
\usepackage{titlesec}
\usepackage{xcolor}

\definecolor{SlopeNavy}{HTML}{17324D}
\definecolor{SlopeBlue}{HTML}{2C6EAA}
\definecolor{cadencecol}{HTML}{EEF1F4}

\titleformat{\section}
  {\Large\bfseries\color{SlopeNavy}}
  {\thesection}
  {0.75em}
  {}

\titleformat{\subsection}
  {\large\bfseries\color{SlopeBlue}}
  {\thesubsection}
  {0.75em}
  {}

\usepackage{etoolbox}
\makeatletter
\patchcmd{\@maketitle}
  {\@title}
  {\begin{minipage}{\dimexpr\linewidth-2\leftmargini\relax}\centering\@title\end{minipage}}
  {}{}
\patchcmd{\@maketitle}
  {\end{tabular}\par}
  {\end{tabular}\par\vskip 0.4em{\normalsize Slope}\par}
  {}{}
\patchcmd{\abstract}{\quotation}{\quote}{}{}
\patchcmd{\endabstract}{\endquotation}{\endquote}{}{}
\makeatother

\makeatletter
\@ifpackageloaded{caption}{}{\usepackage{caption}}
\AtBeginDocument{%
\ifdefined\contentsname
  \renewcommand*\contentsname{Table of contents}
\else
  \newcommand\contentsname{Table of contents}
\fi
\ifdefined\listfigurename
  \renewcommand*\listfigurename{List of Figures}
\else
  \newcommand\listfigurename{List of Figures}
\fi
\ifdefined\listtablename
  \renewcommand*\listtablename{List of Tables}
\else
  \newcommand\listtablename{List of Tables}
\fi
\ifdefined\figurename
  \renewcommand*\figurename{Figure}
\else
  \newcommand\figurename{Figure}
\fi
\ifdefined\tablename
  \renewcommand*\tablename{Table}
\else
  \newcommand\tablename{Table}
\fi
}
\@ifpackageloaded{float}{}{\usepackage{float}}
\floatstyle{ruled}
\@ifundefined{c@chapter}{\newfloat{codelisting}{h}{lop}}{\newfloat{codelisting}{h}{lop}[chapter]}
\floatname{codelisting}{Listing}

\makeatother
\makeatletter
\@ifpackageloaded{caption}{}{\usepackage{caption}}
\@ifpackageloaded{subcaption}{}{\usepackage{subcaption}}
\makeatother
\usepackage{bookmark}
\IfFileExists{xurl.sty}{\usepackage{xurl}}{} 
\makeatletter
\@ifundefined{xmpquote}{\newcommand{\xmpquote}[1]{#1}}{}
\makeatother
\hypersetup{
  pdftitle={CADENCE: Distributional Next-Transaction Forecasting for Business Cash-Flow Underwriting},
  pdfauthor={Alexander Wu; Jingchao Huang},
  pdfkeywords={\xmpquote{cash-flow
underwriting}, \xmpquote{small-business lending}, \xmpquote{next-event
prediction}, \xmpquote{probabilistic forecasting}, \xmpquote{temporal
point processes}, \xmpquote{transformers}, \xmpquote{transaction data}},
  colorlinks=true,
  linkcolor={blue},
  filecolor={Maroon},
  citecolor={Blue},
  urlcolor={Blue},
  pdfcreator={LaTeX via pandoc}}

\title{CADENCE: Distributional Next-Transaction Forecasting for Business
Cash-Flow Underwriting}
\author{Alexander Wu \and Jingchao Huang}
\date{}
\begin{document}
\maketitle
\begin{abstract}
Recurring cash flows are the backbone of cash-flow underwriting: they
separate durable revenue from one-off deposits, expose standing
obligations, and ultimately determine ability to pay. For small
businesses, recurrence is both indispensable and hard to estimate,
because much of their income arrives as ambiguous transfers, most
payment streams are short, and schedules follow calendar conventions
that frequency-domain methods treat as noise. We present CADENCE,
Calendar-Aware Distributional Event Network for Cash-Flow Estimation, a
compact causal Transformer that treats each transaction as a token and
predicts the amount and a full, nonparametric timing distribution of
every cash flow's next payment. Inheriting transaction semantics from an
external labeler and a frozen text encoder keeps CADENCE compact yet
general across the long tail of small-business counterparties, while
predicting a calibrated distribution rather than an opaque embedding
makes its outputs interpretable enough to feed existing underwriting
models and power credit decisions. Ablations confirm the role of each
design choice, with counterparty semantics driving recall on short
histories, calendar inputs driving timing accuracy, and no-event
supervision driving precision. Against the production FFT baseline,
CADENCE improves one-day timing accuracy by 93.3\% and next-event F1 by
6.7\%, while producing calibrated distributions that serve directly as a
confidence gate for downstream decisions. In an end-to-end credit
evaluation on production applications, replacing FFT recurrence features
with CADENCE features raises KS by 4.8\%, raises the approval rate at a
fixed loss rate by 18.2\%, and lowers the loss rate at a fixed approval
rate by 13.1\%.
\end{abstract}

\section{Introduction}\label{sec-introduction}

\subsection{The recurrence problem in business cash-flow
underwriting}\label{sec-cash-flow-problem}

Open banking has made the bank account one of the richest sources of
credit data a lender can obtain. Unlike a bureau file, which records how
a borrower has handled credit in the past, a bank account shows the live
pulse of its finances, and cash-flow events such as an income drop or
spending surge often precede the missed payments that bureaus eventually
report. Consumer lending has embraced this signal, and cash-flow scores
now routinely complement traditional credit scores. Small-business
lending should, in principle, benefit even more. Small businesses
account for roughly 43.5\% of U.S. GDP (U.S. Small Business
Administration Office of Advocacy 2024), working capital is existential
to their operations in a way it rarely is for consumers, bank statements
are already a standard part of most business loan applications, and
credit bureau coverage is notoriously thin for small businesses. Yet
cash-flow underwriting for small businesses remains largely unsolved,
not for lack of demand, but for tractability: business cash flows are
simply hard to understand. This paper addresses one of the central
pieces of that problem: recognizing and forecasting the cash flows that
recur.

A cash flow stream, meaning all transactions between one borrower and
one counterparty in one direction, is recurring when it repeats on a
stable schedule. Identifying such streams is valuable in every form of
cash-flow underwriting to the extent that underwriting is largely a
question of ability to pay: recurring inflows help determine future
income, and recurring outflows help determine future obligations such as
payroll, rent, and debt servicing. Once a payment is expected, even its
absence becomes informative, whether it signals a missed paycheck, a
lost customer, or a skipped installment.

For consumers, though, recurrence classification is seldom the
bottleneck to underwriting. A typical individual receives a single
biweekly paycheck identifiable by static keywords (e.g.,
\texttt{DIRECT\ DEP}, \texttt{ADP}, \texttt{GUSTO}), spends within a
narrow set of categories that can be captured by a standard taxonomy
(e.g., rent, groceries, utilities), and transacts with a concentrated
set of counterparties that can be captured by a short list (e.g.,
Amazon, Costco, Netflix). As a result, transaction-level tagging alone
is often sufficient for computing income, debt, and expenses: all the
inputs to ability to pay.

Businesses offer no such shortcut, because their cash flows are complex,
diverse, and context-dependent (Wu 2025). \textbf{Complexity:} a
business has many more distinct cash flows than a consumer, and each is
more complex along every dimension, spanning many counterparties and
channels, from point-of-sale settlements and platform payouts to
financing, owner capital, and transfers among its own accounts, and
fluctuates over time with the seasons and demand. \textbf{Diversity:}
cash-flow patterns depend heavily on industry. A restaurant, a roofing
contractor, and an e-commerce seller have entirely different income,
financing, and expense profiles, and no single taxonomy fits them all.
\textbf{Context-dependence:} much of business cash flow is locally
ambiguous and cannot be interpreted from individual transactions. In our
corpus, roughly a third of inflow dollars arrive as generic transfers,
and for one business in three, such transfers make up at least half of
all inflows. Whether a monthly Zelle payment is a client's retainer, a
loan, or an owner infusion is often ambiguous from the descriptor alone,
and in industries such as services and wholesale trade, the entirety of
revenue can arrive this way. Such ambiguity yields only to temporal
context: the rhythm, regularity, and persistence of a counterparty's
payments over time.

At Slope, recurrence has become the backbone of underwriting: features
built on recurring cash flows carry over 50\% of the feature influence
in our probability-of-default model and significant weight in
credit-line assignment, so small errors in recurrence travel directly
into approvals, limits, and losses. Our current production system, like
many periodicity detectors, estimates that recurrence in the frequency
domain. A fast Fourier transform (FFT) of each stream's event series
yields a dominant period together with scores such as regularity and
signal strength, and these signals feed the downstream credit models
(Cooley and Tukey 1965). The approach is transparent, inexpensive, and
effective on long, regular series. However, it has several shortfalls:

\begin{itemize}
\tightlist
\item
  \textbf{Needs long histories.} FFT treats a sparse sequence of
  payments as a uniformly sampled signal whose frequency resolution
  grows with the length of the observed window, so it needs several
  cycles to determine periodicity. This is a poor fit for a long-tailed
  corpus in which most streams have only a handful of events and where
  new cash flows matter most precisely when their history is shortest.
\item
  \textbf{Calendar blindness.} FFT decomposes a signal into sinusoids
  with fixed periods in days, but calendar schedules are not fixed in
  days: months run 28 to 31 days and payments shift around weekends and
  holidays, so a semi-monthly payroll on the 1st and 15th smears into a
  noisy two-week cycle, and a payment that alternates between accounts
  with different posting delays looks irregular.
\item
  \textbf{No semantics.} FFT cannot see the transaction descriptor or
  borrow strength from similar counterparties across the portfolio, such
  as a known lender or supplier.
\item
  \textbf{Backward-looking.} Most fundamentally, FFT assumes the
  frequency content of a signal is fixed over its history, but business
  cash flows are non-stationary: cash flows start, stop, and change
  frequency continuously. FFT summarizes the average past rather than
  predicting when, or whether, the next payment will arrive, and with
  what confidence.
\end{itemize}

These shortfalls are material enough to cause issues in production at
scale, where they surface as underestimated revenue, missed new revenue
streams, and missed significant events.

\subsection{Approach and contributions}\label{sec-contributions}

\begin{figure}[H]

\centering{

\includegraphics[width=1\linewidth,height=\textheight,keepaspectratio]{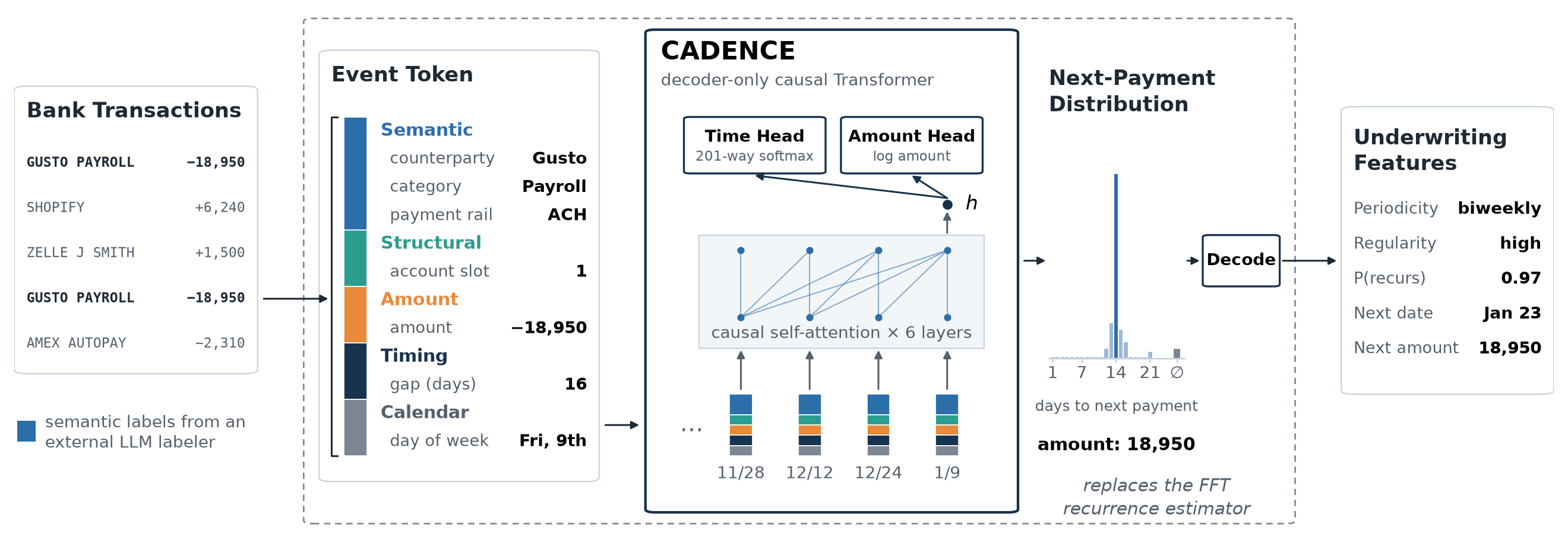}

}

\caption{\label{fig-system-overview}CADENCE end to end for one example
payroll stream. Each transaction becomes one event token that combines
LLM-derived semantic labels with structural, amount, timing, and
calendar fields. A causal Transformer reads the stream's tokens, and
from the final hidden state a time head predicts a calibrated
distribution over the days to the next payment, including no event,
while an amount head predicts its size. Decoding the distribution yields
the underwriting features. The dashed region replaces the FFT recurrence
estimator.}

\end{figure}

We replace frequency-domain recurrence estimation with CADENCE, a
calendar-aware next-event model that borrows its framing from language
modeling (Figure~\ref{fig-system-overview}). Each transaction in a
stream becomes a token, and a compact causal Transformer (Vaswani et al.
2017) reads the tokens so far to predict a full probability distribution
over the number of days until the next one, up to 200 days, together
with an explicit outcome that no further event occurs and an estimate of
the next amount. Every quantity underwriting needs, from whether a
stream is live to when it pays next and how sure the model is, becomes a
reading of that distribution, so downstream credit models need no
redesign.

The design reflects the constraints of business lending rather than
those of card networks or large consumer banks. Small-business
portfolios contain tens of millions of transactions rather than tens of
billions, and the model must be fast and simple to serve inside live
underwriting. Instead of learning what transactions mean from raw
descriptions, CADENCE inherits that knowledge from Slope's LLM
transaction labeler and a frozen text encoder, which lets a
2.7-million-parameter model train in about 30 GPU-hours and still
generalize across a long tail of counterparties.

Our contributions are as follows:

\begin{itemize}
\tightlist
\item
  \textbf{Interpretable, nonparametric output.} We frame recurrence as
  stream-level next-event prediction with explicit termination, so the
  model's output is a calibrated probability distribution over
  observable outcomes rather than an opaque embedding. Because the
  distribution is nonparametric, it follows the skewed, long-tailed, and
  multi-peaked timing of real cash flows, exposes calendar structure
  such as weekend shifts, four-or-five-week months, and day-of-week
  anchoring, and directly yields the signals required for underwriting:
  regularity, periodicity, signal strength, and the probability that a
  cash flow recurs.
\item
  \textbf{Prior-driven tokenization.} Each transaction becomes a single
  token built from LLM-derived counterparty, category, and payment-rail
  semantics, amount, inter-arrival time, and calendar position, which
  makes the model data-efficient enough to train at business-lending
  scale.
\item
  \textbf{Mechanisms.} Ablations attribute exact-day timing to calendar
  inputs, short-history recall to counterparty semantics, and precision
  to explicit no-event supervision.
\item
  \textbf{Direct evidence.} Against the production FFT baseline, CADENCE
  improves one-day timing accuracy by 93.3\% and next-event F1 by 6.7\%;
  on streams with three to five events, next-event recall rises by
  510\%.
\item
  \textbf{Credit impact.} Replacing FFT recurrence features with CADENCE
  features in a credit-application evaluation raises KS by 4.8\%, raises
  the approval rate at a fixed loss rate by 18.2\%, and lowers the loss
  rate at a fixed approval rate by 13.1\%.
\end{itemize}

CADENCE is a feature-generation component: it improves the recurrence
inputs to underwriting and leaves credit policy and final decisions to
the full underwriting stack.

\section{Related Work}\label{sec-related-work}

\subsection{Periodicity detection and event-time
models}\label{sec-related-periodicity}

Production recurrence detection has historically treated recurrence as a
property of spacing. Rule-based systems classify a series as recurring
when amounts fall within a tolerance and intervals match a known
cadence, while spectral methods such as FFT (Cooley and Tukey 1965)
estimate the dominant period directly and score how much of the series
it explains. Both are interpretable and cheap, and both share the
shortfalls described in Section~\ref{sec-cash-flow-problem}: they need
several cycles of history, treat calendar-anchored schedules as noisy
fixed intervals, and cannot share evidence across counterparties.
Slope's production baseline is a spectral estimator of this kind.

Temporal point processes supply the statistical framing for our task:
they model the distribution of the next event time given the history.
Neural point processes encode that history with recurrent or
attention-based networks (Du et al. 2016; Zuo et al. 2020), and
intensity-free formulations model the inter-event density directly,
including multimodal mixtures (Shchur et al. 2020). CADENCE's time head
is a discrete-time instance of the same idea: because transactions post
at daily resolution, it predicts a categorical distribution over days
plus a no-event class, which captures multimodality and termination
without parametric assumptions.

\subsection{Transaction foundation models}\label{sec-related-foundation}

A parallel line of work at large financial institutions has shown that
Transformers trained on raw transaction histories learn broadly useful
representations. PRAGMA pre-trains a family of encoder-only models, from
10 million to 1 billion parameters, with masked modeling on 24 billion
consumer banking events from 26 million users, and adapts the shared
backbone to credit scoring, fraud detection, lifetime value, and other
tasks (Ostroukhov et al. 2026). Each event is decomposed into
key-value-time tokens, and free-text fields such as transaction
descriptions are split into subword tokens whose embeddings are learned
from scratch. Recurring-transaction detection appears among PRAGMA's
downstream tasks as a binary prediction of whether a subscription
payment repeats in the following month, and the improvement there is
modest, nearly flat across a hundredfold increase in model size, and
essentially unchanged by adding a pre-trained text encoder.

TREASURE, Visa's payment foundation model, comes closest to our
formulation (Yeh et al. 2026). It is a decoder-only Transformer trained
on six billion card transactions to predict the attributes of each
cardholder's next transaction, with log-normal distributions for
numerical attributes and learned embedding tables for high-cardinality
attributes such as merchant identity. Nubank, Visa, and Featurespace
have likewise pre-trained autoregressive models on consumer transaction
sequences for downstream fine-tuning ({Braithwaite et al.} 2025; {Dou et
al.} 2025; Skalski et al. 2023).

\subsection{Gaps for business cash-flow
recurrence}\label{sec-related-gaps}

These systems demonstrate that next-transaction modeling works at scale,
but several gaps remain for business cash-flow recurrence:

\begin{itemize}
\tightlist
\item
  \textbf{Data regime.} They learn from billions of consumer or card
  events, while business lending offers orders of magnitude fewer
  transactions spread over a long tail of counterparties.
\item
  \textbf{Granularity.} PRAGMA answers a binary monthly question and
  TREASURE predicts a card's next transaction across all merchants, so
  neither forecasts when a specific counterparty will pay next, how
  much, or whether it has stopped.
\item
  \textbf{Shape of uncertainty.} A single log-normal, like a single
  dominant period, imposes one parametric shape on payment timing that
  is often skewed, long-tailed, or split between competing calendar
  anchors, such as the 15th versus the last business day of the month.
\item
  \textbf{Semantics.} Learning text or merchant embeddings from scratch
  bounds a model's knowledge by the institution's own data, and both
  Revolut and Visa report gains from adding pre-trained language
  representations (Ostroukhov et al. 2026; Fan et al. 2025).
\item
  \textbf{Interpretability.} General-purpose embeddings must be adapted
  before use and are opaque to credit policy, whereas underwriting needs
  outputs with direct meaning, such as the probability that a payment
  arrives.
\end{itemize}

Section~\ref{sec-design-principles} describes how CADENCE addresses
these gaps.

\section{Methodology}\label{sec-methodology}

\subsection{Problem formulation}\label{sec-problem-formulation}

The unit of prediction is a cash flow stream: all transactions between
one business and one counterparty in one direction. Inflows and outflows
with the same counterparty form separate streams, because a customer who
both pays a business and receives refunds from it produces two different
cash-flow patterns. Streams are built from the counterparty extracted by
Slope's transaction labeler rather than from raw descriptors, which
embed dates, reference numbers, and processor codes that would otherwise
fragment a single relationship into many apparent counterparties.

A stream is an ordered sequence of events \(e_1, \dots, e_n\) with
posting dates \(t_1 \le \dots \le t_n\). Given the history up to
\(e_n\), the model predicts the gap to the next event,

\[
y = t_{n+1} - t_n \in \{1, \dots, 200\} \cup \{\varnothing\},
\]

where \(\varnothing\) denotes that no further event occurs within 200
days, together with the amount of the next event. The output is
therefore a distribution \(p(y \mid e_{1:n})\) rather than a single
date, and every quantity used downstream, from event probability and
most likely date to confidence, is a summary of it.

The horizon is a modeling choice with real consequences. It must be long
enough that the no-event class means a stream has stopped rather than
that it pays on a slow schedule, yet short enough to keep the output
compact and to confirm terminations within a reasonable observation
window. Two hundred days spans cadences from daily to semiannual and
clears the quarter by a wide margin; Section~\ref{sec-horizon-results}
shows that a 90-day horizon, which falls inside the 89-to-92-day span of
three calendar months, pushes most quarterly payments into the no-event
class.

\subsection{Design principles}\label{sec-design-principles}

CADENCE is shaped by the constraints of business lending. Data is
limited: business lending generates tens of millions of transactions,
not billions, which rules out the pre-training budgets of consumer
foundation models. Simultaneously, those transactions are far more
heterogeneous than consumer data, spread across a longer tail of
counterparties, categories, and payment channels. And the model must be
fast to serve inside live underwriting, with outputs that remain
interpretable to credit models and analysts. Three design principles
follow:

\textbf{We inherit semantics instead of learning them.} Frequency-domain
methods use only the timing of payments and discard their descriptors,
while PRAGMA and TREASURE learn semantics end to end from subword tokens
or merchant identifiers, which bounds their knowledge by the
institution's own data. CADENCE instead inherits semantics from two
pre-trained components: Slope TransFormer, a fine-tuned LLM transaction
labeler, maps each raw descriptor to a canonical counterparty, category,
subcategory, and payment rail (Wu 2023), and a frozen text encoder
embeds the counterparty name (Xiao et al. 2023). The sequence model
never reads raw descriptors; it consumes only these labels and
embeddings. The language model contributes world knowledge from
pre-training, such as that Gusto and ADP are payroll processors or that
a revenue-based financing debit repeats until the advance is repaid,
while the sequence model learns only how such flows behave over time.
Inherited semantics bring two practical benefits: first, a new
counterparty name embeds near semantically similar ones, so the model
generalizes to counterparties it has never seen; second, a labeled
counterparty is informative from the first payment, so fewer payments
are needed to recognize a new cash flow stream as recurring
(Section~\ref{sec-ablation-results}).

\textbf{We encode the calendar in addition to spacing.} Business
payments are scheduled by calendar position, such as every other Friday,
the 1st and 15th, or the last business day of the month, and they shift
around weekends, holidays, and posting delays. FFT represents a schedule
as fixed-period sinusoids, so these shifts read as noise, and foundation
models take time as an input but predict next-event timing as a single
log-normal or not at all. CADENCE reads each payment's day of week, day
of month, and month-end and quarter-end position alongside the elapsed
gap, separates schedules that post to different accounts with an
account-slot input, and resolves its output to the day.

\textbf{We predict an interpretable, nonparametric distribution.} Where
FFT offers no explicit prediction and foundation models produce
general-purpose embeddings meant to be fine-tuned for downstream
applications, CADENCE predicts a calibrated distribution over the days
until the next payment, including an explicit no-event outcome
(Section~\ref{sec-causal-model}). It makes no assumption about that
distribution's shape: the output is a categorical over the next 200
days, whereas FFT assumes a single dominant period and TREASURE a single
log-normal. The prediction can therefore model the messy, often
non-Gaussian structure of real cash-flow cadences, which may be skewed,
long-tailed, or multi-peaked (Figure~\ref{fig-distribution-gallery}).
That shape carries information beyond the most likely date: peak width
reflects confidence, multiple peaks reflect overlapping cadences such as
biweekly payroll with a monthly bonus, and the no-event probability
reflects termination risk. And because every value is the probability of
an observable event, the output can be audited, thresholded, and
monitored directly (Figure~\ref{fig-calibration}).

Table~\ref{tbl-related-comparison} summarizes these choices against both
alternatives.

\begin{longtable}[]{@{}
  >{\raggedright\arraybackslash}p{(\linewidth - 8\tabcolsep) * \real{0.1600}}
  >{\raggedright\arraybackslash}p{(\linewidth - 8\tabcolsep) * \real{0.1800}}
  >{\raggedright\arraybackslash}p{(\linewidth - 8\tabcolsep) * \real{0.2200}}
  >{\raggedright\arraybackslash}p{(\linewidth - 8\tabcolsep) * \real{0.2200}}
  >{\raggedright\arraybackslash}p{(\linewidth - 8\tabcolsep) * \real{0.2200}}@{}}
\caption{CADENCE compared with the FFT baseline and transaction
foundation models.}\label{tbl-related-comparison}\tabularnewline
\toprule\noalign{}
\begin{minipage}[b]{\linewidth}\raggedright
\end{minipage} & \begin{minipage}[b]{\linewidth}\raggedright
FFT baseline
\end{minipage} & \begin{minipage}[b]{\linewidth}\raggedright
PRAGMA
\end{minipage} & \begin{minipage}[b]{\linewidth}\raggedright
TREASURE
\end{minipage} & \begin{minipage}[b]{\linewidth}\raggedright
\cellcolor{cadencecol}\textbf{CADENCE}
\end{minipage} \\
\midrule\noalign{}
\endfirsthead
\toprule\noalign{}
\begin{minipage}[b]{\linewidth}\raggedright
\end{minipage} & \begin{minipage}[b]{\linewidth}\raggedright
FFT baseline
\end{minipage} & \begin{minipage}[b]{\linewidth}\raggedright
PRAGMA
\end{minipage} & \begin{minipage}[b]{\linewidth}\raggedright
TREASURE
\end{minipage} & \begin{minipage}[b]{\linewidth}\raggedright
\cellcolor{cadencecol}\textbf{CADENCE}
\end{minipage} \\
\midrule\noalign{}
\endhead
\bottomrule\noalign{}
\endlastfoot
Data & None (fit per stream) & 24B consumer banking events & 6B card
transactions & \cellcolor{cadencecol}47M business bank transactions \\
Objective & Spectral period estimate & Masked modeling (encoder) & Next
transaction (decoder) & \cellcolor{cadencecol}Next event per stream
(decoder) \\
Semantics & None & Subword embeddings learned from scratch & Learned
merchant embeddings & \cellcolor{cadencecol}LLM labels and frozen text
embeddings \\
Calendar & None; fixed-period sinusoids & Timestamps as tokens & Not
reported & \cellcolor{cadencecol}Calendar inputs; day-resolution
output \\
Output & Period and strength scores & Embedding; binary
repeat-next-month task & Embedding; next card transaction, any merchant
& \cellcolor{cadencecol}Calibrated next-payment distribution and
amount \\
Timing uncertainty & None & Not predicted & Single log-normal &
\cellcolor{cadencecol}Categorical over days, with no-event class \\
Model size & Closed form & 10M--1B parameters & Not reported &
\cellcolor{cadencecol}2.7M parameters \\
\end{longtable}

\subsection{Transactions as tokens}\label{sec-event-representation}

\begin{figure}[H]

\centering{

\includegraphics[width=1\linewidth,height=\textheight,keepaspectratio]{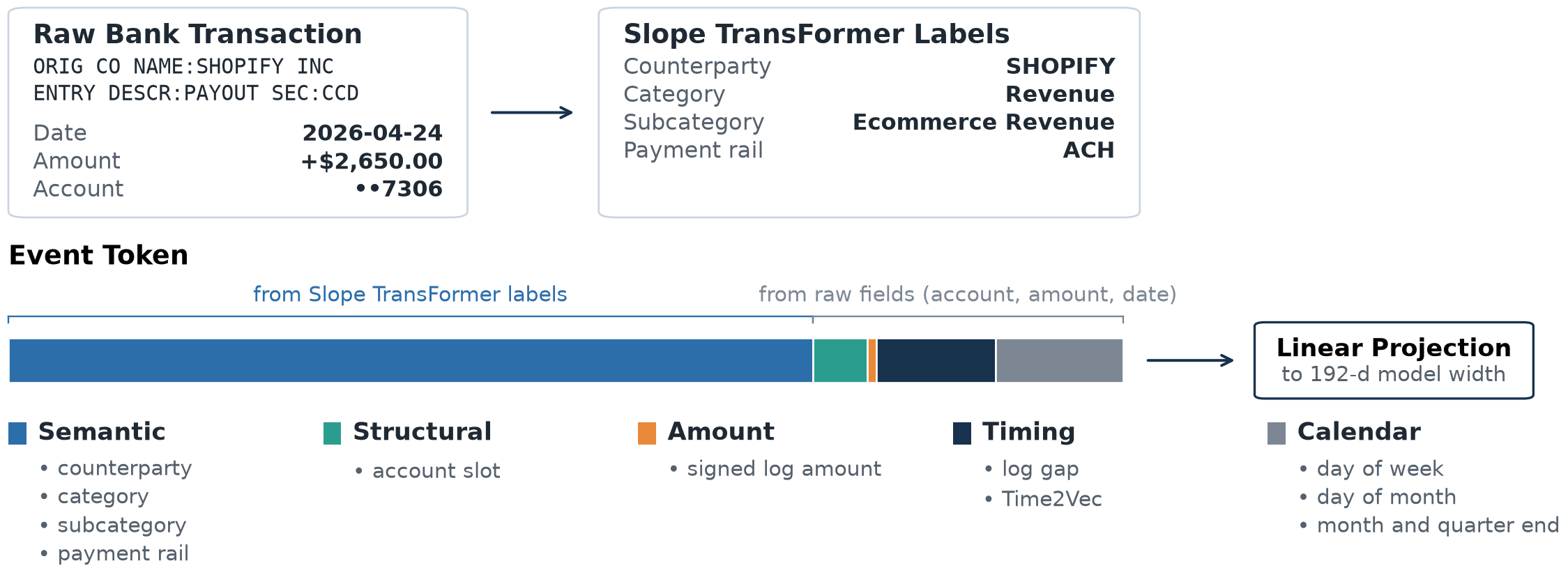}

}

\caption{\label{fig-tokenization}Transactions as tokens. Slope
TransFormer converts a raw bank descriptor into a canonical
counterparty, category, subcategory, and payment rail. Each transaction
then becomes a single event token: semantic inputs come from the labels,
while account, amount, timing, and calendar inputs come from the raw
transaction fields. A linear layer projects the token to the model width
of 192.}

\end{figure}

Each transaction becomes exactly one token
(Figure~\ref{fig-tokenization}). This is a deliberate contrast with
key-value-time tokenization, in which a single event expands into many
field and subword tokens (Ostroukhov et al. 2026). With one token per
event, the context window is measured in transactions: 128 positions
cover a decade of monthly payments, more than two years of weekly ones,
or several months of daily settlements. The token concatenates five
groups of inputs and projects them linearly to the model width of 192
(Table~\ref{tbl-event-token}).

Semantic inputs describe what the transaction is. The counterparty name
is encoded once per stream by the frozen BGE-small encoder (Xiao et al.
2023) and compressed by a learned projection, while category,
subcategory, and payment rail receive small learned embeddings. The
account slot identifies which of the customer's accounts, up to eight,
the transaction posted to, so the model can recognize a schedule that
alternates between accounts as one pattern rather than as noise. The
amount enters as a signed logarithm, which preserves direction and
compresses the several orders of magnitude between a software
subscription and a wholesale invoice. Timing enters as the logarithm of
the gap since the previous event, both directly and through a Time2Vec
encoding that lets the model learn periodic functions of the gap (Kazemi
et al. 2019). Calendar inputs encode the day of week, the day of month,
and month-end and quarter-end flags. The gap tells the model how long it
has been since the last payment, while the calendar tells it where the
stream now stands, and schedules such as the 1st and 15th or the last
business day of the month are defined by the latter.

\begin{longtable}[]{@{}
  >{\raggedright\arraybackslash}p{(\linewidth - 4\tabcolsep) * \real{0.3333}}
  >{\raggedright\arraybackslash}p{(\linewidth - 4\tabcolsep) * \real{0.3333}}
  >{\raggedright\arraybackslash}p{(\linewidth - 4\tabcolsep) * \real{0.3333}}@{}}
\caption{Components of an event
token.}\label{tbl-event-token}\tabularnewline
\toprule\noalign{}
\begin{minipage}[b]{\linewidth}\raggedright
Group
\end{minipage} & \begin{minipage}[b]{\linewidth}\raggedright
Input
\end{minipage} & \begin{minipage}[b]{\linewidth}\raggedright
Encoding
\end{minipage} \\
\midrule\noalign{}
\endfirsthead
\toprule\noalign{}
\begin{minipage}[b]{\linewidth}\raggedright
Group
\end{minipage} & \begin{minipage}[b]{\linewidth}\raggedright
Input
\end{minipage} & \begin{minipage}[b]{\linewidth}\raggedright
Encoding
\end{minipage} \\
\midrule\noalign{}
\endhead
\bottomrule\noalign{}
\endlastfoot
Semantic & Counterparty (shared across the stream) & Frozen BGE-small
with learned projection \\
& Category, subcategory, payment rail & Learned embeddings \\
Structural & Account slot & Learned embedding \\
Amount & Signed amount & Signed logarithm \\
Timing & Gap since previous event & Log days and Time2Vec \\
Calendar & Day of week, day of month & Learned embeddings \\
& Month end, quarter end & Binary flags \\
Token & Concatenation & Linear projection to model width \\
\end{longtable}

\subsection{Causal sequence model and distributional
heads}\label{sec-causal-model}

A decoder-only Transformer (Vaswani et al. 2017), a stack of
self-attention layers with a causal mask, reads up to the 128 most
recent event tokens of a stream. The model has six layers, four
attention heads, a width of 192, and a 768-dimensional feed-forward
block with GELU activations, for 2.7 million parameters in total. The
causal mask restricts each position to itself and earlier events, so the
representation at the last position summarizes exactly the history
available at prediction time. We use no positional encodings, a choice
TREASURE also makes (Yeh et al. 2026). Order is implied by the causal
mask, and temporal position is carried more faithfully by the gap and
calendar inputs than by an integer index, since two streams with the
same number of events can span weeks or years.

Two heads read the final hidden state. The time head is a
layer-normalized linear map to 201 logits, one for each day from 1 to
200 and one for no event, followed by a softmax. The amount head
regresses the logarithm of the next amount, which yields a stable point
estimate.

The categorical time head is the central design choice. A regression
head returns a single expected gap, and a log-normal head a single mode
with a spread; neither can say that a payment will arrive either four or
five weeks from now, or on either the 15th or the last business day of
the month. A softmax over days is nonparametric: it represents any such
shape, places termination on the same probability scale as timing, and
matches the daily resolution at which transactions post. It also turns
every downstream quantity into a simple reading of the distribution: the
probability that an event occurs at all, the most likely day, the
probability mass near that day, and the entropy of the whole
distribution. Section~\ref{sec-distribution-results} shows that these
shapes are interpretable and that the confidence they express tracks
realized accuracy.

\begin{figure}[H]

\centering{

\includegraphics[width=1\linewidth,height=\textheight,keepaspectratio]{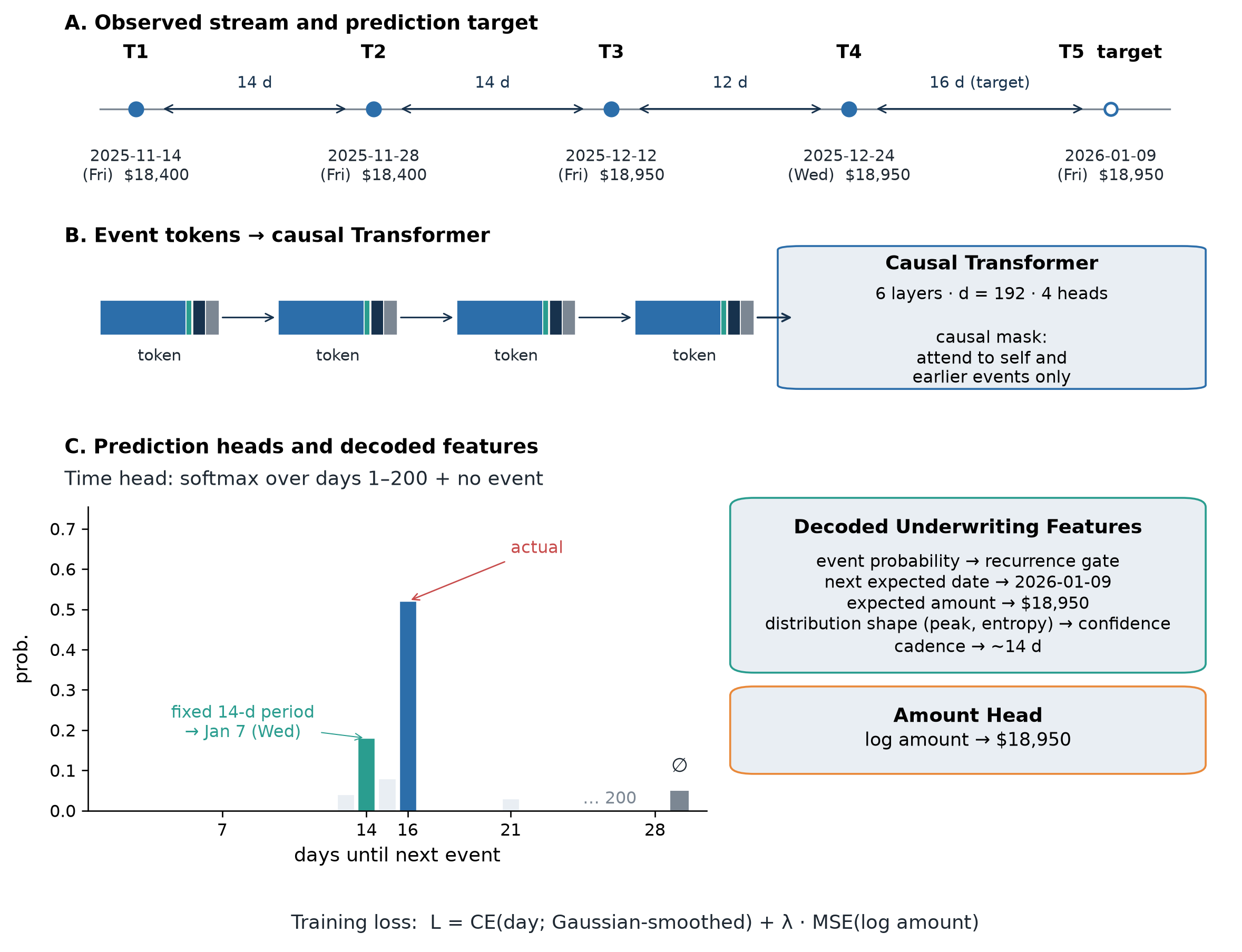}

}

\caption{\label{fig-architecture}Next-event prediction for one stream,
illustrated with a schematic biweekly Friday payroll whose December
payment was made early, on a Wednesday, before the holiday. Each past
payment becomes a token, a causal Transformer summarizes the history at
the last event, and two heads predict a distribution over the days until
the next event, including an explicit no-event class, and the next
amount. Because the model sees calendar position, it expects the
schedule to return to Friday 16 days later rather than repeat a fixed
14-day interval that would land on a Wednesday. The distribution is then
decoded into the recurrence features used in underwriting.}

\end{figure}%

\subsection{Training}\label{sec-training-objective}

Training follows the autoregressive recipe of language models. Every
prefix of a stream with at least three events is an example whose target
is the event that follows it, so a stream with \(n\) events contributes
\(n - 3\) examples. When the observed data extend at least 200 days
beyond a stream's last event without another event, that final prefix
becomes a terminal example labeled \(\varnothing\). Terminal examples
are the only supervision the model receives about streams ending, and
Section~\ref{sec-ablation-results} shows that they are essential for
precision. Figure~\ref{fig-architecture} illustrates a single example.

The time head is trained with cross-entropy against a Gaussian-smoothed
target centered on the true gap, while terminal examples place all
target mass on the no-event class. Smoothing rewards near misses,
reflecting that ACH settlement and posting conventions routinely move a
payment by a day, and it spreads gradient to neighboring days so that
rare gaps still shape the distribution. The amount head is trained with
squared error on the log amount, masked for terminal examples and
down-weighted relative to the time loss so that timing dominates the
shared representation:

\[
\mathcal{L} = -\sum_{k=1}^{201} \tilde{q}_k \log p_k + \lambda\,\bigl(\log \hat{a} - \log a\bigr)^2,
\]

where \(\tilde{q}\) is the smoothed target distribution over the 201
outcomes, \(p\) is the predicted distribution, \(a\) and \(\hat{a}\) are
the actual and predicted amounts, and \(\lambda < 1\) weights the amount
loss.

CADENCE is trained on about 1.3 billion next-event predictions in total,
which takes about 30 GPU-hours on a single GPU.

\subsection{Decoding and production
integration}\label{sec-production-decoding}

The predicted distribution is decoded into the signals required for
underwriting, so CADENCE can replace FFT without changes to the
downstream credit models. The most likely day gives the next expected
date and the stream's cadence, and the amount head gives the expected
next amount. Whether a stream counts as recurring is decided by a
recurrence gate. In its simplest form, the gate thresholds the event
probability \(1 - p(\varnothing)\). However, we find that a small
logistic model over features of the distribution, such as event
probability, peak sharpness, and entropy, outperforms a gate based on
event probability alone, demonstrating the value of the distributional
prediction.

CADENCE runs as a feature-generation component inside the underwriting
pipeline: it consumes the same labeled transactions as FFT and emits the
signals required for underwriting in the same form. Credit policy,
knockout rules, PD score, and final decisions remain governed by the
existing underwriting stack.

\section{Experimental Setup}\label{sec-evaluation}

We evaluate CADENCE at three levels: direct next-payment accuracy
against the FFT baseline, controlled ablations of individual design
choices, and downstream utility in credit-application decisions.

\subsection{Data}\label{sec-data}

All training and evaluation use roughly 47 million business bank
transactions from Slope's production underwriting data. Every
transaction is first labeled by Slope TransFormer, a fine-tuned LLM
transaction labeler, with a canonical counterparty, a category and
subcategory, and, where identifiable, a payment rail such as ACH, check,
Zelle, card, or wire (Wu 2023). Customers are eligible when they have at
least 180 days of history, and streams are modeled when they contain at
least three events.

The data are long-tailed along every axis that matters for recurrence.
Most streams are short while a few contain hundreds of events, so the
same parameters must handle near-daily card settlements and a handful of
quarterly tax payments. Most counterparties likewise appear in only a
few streams, the regime in which embeddings learned from scratch are
weakest.

Customers are split into training, validation, and holdout partitions by
stable hashing of the customer identifier, so no customer's history
appears in more than one partition. Each validation and holdout customer
receives a reference date that plays the role of the underwriting
decision date: the model observes only transactions on or before that
date, and the first transaction of each stream after it serves as the
evaluation outcome. All reported direct metrics use the validation
partition.

\subsection{Direct evaluation}\label{sec-next-payment-eval}

The baseline is the production FFT recurrence estimator applied to
exactly the same point-in-time histories, and both methods are scored
against the same future events. Next-payment metrics are computed per
stream against the first event after the reference date. Among streams
with both an emitted prediction and an observed next event within 200
days, timing accuracy is the share of predictions within one and within
three calendar days of the actual date. Next-event precision is the
share of emitted predictions that are followed by an observed event,
next-event recall is the share of observed events that received a
prediction, and F1 combines the two.

Because recurrence metrics trade precision against recall, the
comparison with FFT uses a standardized operating point: the gate
threshold is chosen once so that CADENCE flags streams as recurring with
the same precision as the FFT baseline. Matching precision ensures that
every reported gain comes without admitting more false recurrences than
the production system. This comparison covers validation streams with at
least five prior events. Ablations instead hold a single
event-probability threshold fixed across variants and compare each
variant with the full model, scored on all validation streams with at
least three events so that short histories can be examined separately.
Because absolute accuracy levels are commercially sensitive, all direct
metrics are reported as relative changes, computed as
\((x_{\text{CADENCE}} - x_{\text{FFT}}) / x_{\text{FFT}}\) with the sign
reversed for error metrics, or relative to the full model for ablations.

\subsection{Downstream credit evaluation}\label{sec-downstream-protocol}

The downstream evaluation asks whether better recurrence estimates
produce better credit decisions. It uses labeled credit-application
observations from Slope's production underwriting population. For each
observation, recurrence features are computed point in time from FFT and
from CADENCE, and all other underwriting features are held fixed. Credit
models are gradient-boosted trees (Chen and Guestrin 2016) trained and
scored out of fold across five customer-disjoint folds, with features
selected inside each fold.

We report three metrics that correspond to lending decisions.
Kolmogorov--Smirnov (KS) separation measures how well the risk score
ranks good and bad outcomes overall. The approval rate at a fixed loss
rate measures how many more applicants can be approved without taking
more losses, and the loss rate at a fixed approval rate measures how
many losses are avoided without approving fewer applicants. As with the
direct metrics, results are reported as relative changes from the FFT
feature family.

\section{Results}\label{sec-results}

Headline results are summarized in Table~\ref{tbl-results-overview}, and
detailed results follow in each subsection.

\begin{longtable}[]{@{}lr@{}}
\caption{Headline results relative to the FFT
baseline.}\label{tbl-results-overview}\tabularnewline
\toprule\noalign{}
Result & Relative change \\
\midrule\noalign{}
\endfirsthead
\toprule\noalign{}
Result & Relative change \\
\midrule\noalign{}
\endhead
\bottomrule\noalign{}
\endlastfoot
Timing within one day & +93.3\% \\
Next-event F1 & +6.7\% \\
Credit-application KS & +4.8\% \\
Approval rate at fixed loss rate & +18.2\% \\
Loss rate at fixed approval rate & −13.1\% \\
\end{longtable}

\subsection{Next-payment prediction}\label{sec-next-payment-results}

\begin{figure}[H]

\centering{

\includegraphics[width=0.78\linewidth,height=\textheight,keepaspectratio]{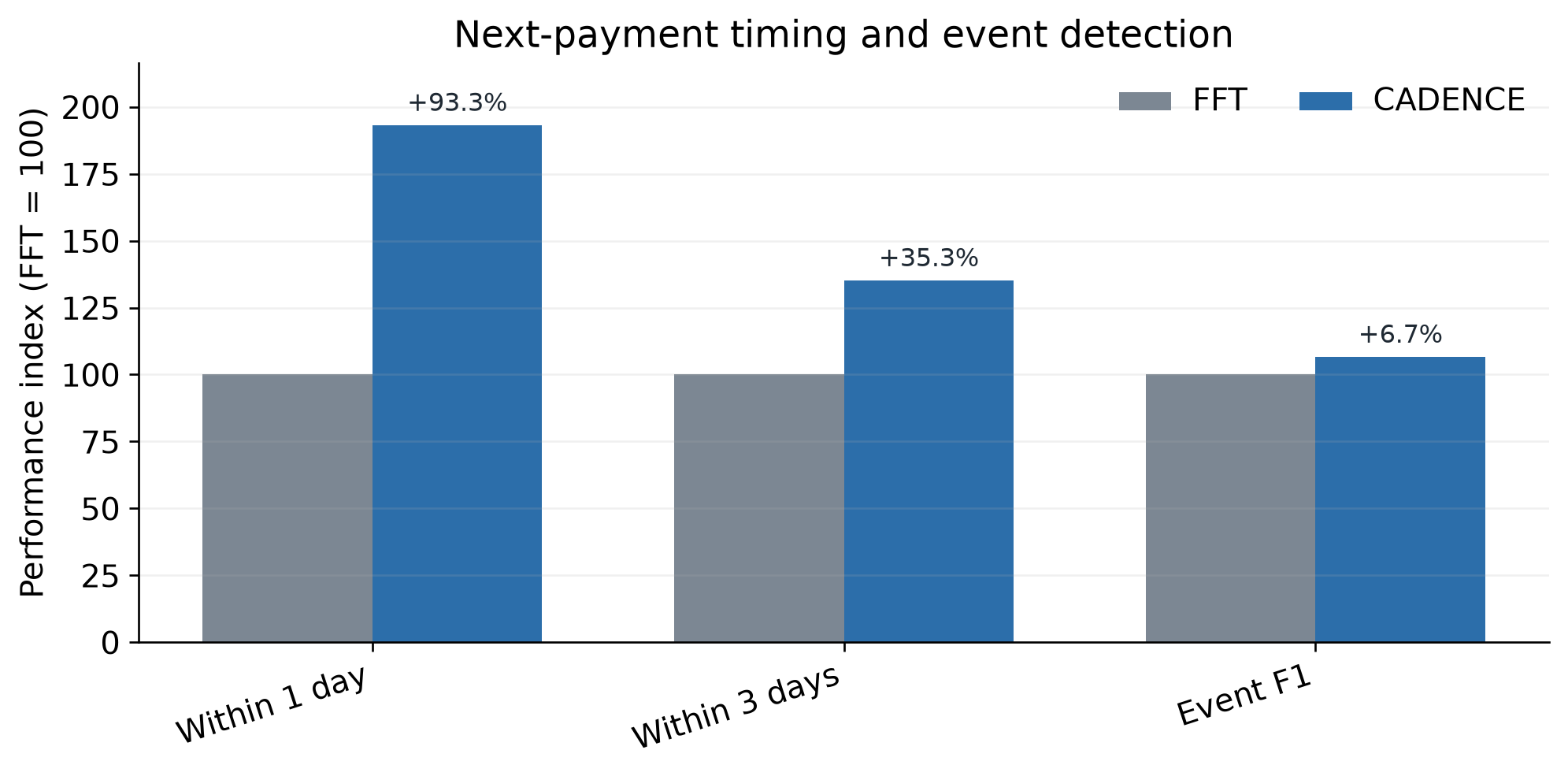}

}

\caption{\label{fig-direct-results}Direct next-payment performance,
normalized so that the FFT baseline equals 100 and higher is better.}

\end{figure}

The largest gain is in exact timing: relative to the FFT baseline,
one-day timing accuracy improves by 93.3\% and three-day accuracy by
35.3\% (Figure~\ref{fig-direct-results}). The larger gain at one day
shows where FFT falls short: it often identifies the right cadence but
the wrong day, landing a few days early or late when a schedule is
anchored to the calendar or shifts around weekends and holidays. Event
detection improves at the same time, with next-event F1 up 6.7\%.

Timing gains hold at every history length, while detection gains
concentrate in short histories (Table~\ref{tbl-history-length}). One-day
timing improves by at least 68\% in every bucket, consistent with
calendar awareness helping any stream anchored to the calendar. On
streams with three to five events, next-event recall rises by 510\% and
precision by 38\%, while on streams with more than 25 events both
methods detect nearly every next payment and the difference narrows to a
few percent. Short streams are where newly recurring revenue and new
obligations first appear, so this is the segment where earlier
recognition matters most for underwriting.

\begin{longtable}[]{@{}
  >{\raggedright\arraybackslash}p{(\linewidth - 6\tabcolsep) * \real{0.2000}}
  >{\raggedleft\arraybackslash}p{(\linewidth - 6\tabcolsep) * \real{0.2667}}
  >{\raggedleft\arraybackslash}p{(\linewidth - 6\tabcolsep) * \real{0.2667}}
  >{\raggedleft\arraybackslash}p{(\linewidth - 6\tabcolsep) * \real{0.2667}}@{}}
\caption{Relative change versus the FFT baseline by history length,
measured at a fixed recurrence
threshold.}\label{tbl-history-length}\tabularnewline
\toprule\noalign{}
\begin{minipage}[b]{\linewidth}\raggedright
Events in history
\end{minipage} & \begin{minipage}[b]{\linewidth}\raggedleft
One-day timing
\end{minipage} & \begin{minipage}[b]{\linewidth}\raggedleft
Next-event precision
\end{minipage} & \begin{minipage}[b]{\linewidth}\raggedleft
Next-event recall
\end{minipage} \\
\midrule\noalign{}
\endfirsthead
\toprule\noalign{}
\begin{minipage}[b]{\linewidth}\raggedright
Events in history
\end{minipage} & \begin{minipage}[b]{\linewidth}\raggedleft
One-day timing
\end{minipage} & \begin{minipage}[b]{\linewidth}\raggedleft
Next-event precision
\end{minipage} & \begin{minipage}[b]{\linewidth}\raggedleft
Next-event recall
\end{minipage} \\
\midrule\noalign{}
\endhead
\bottomrule\noalign{}
\endlastfoot
3--5 & +101\% & +38\% & +510\% \\
6--10 & +79\% & +9\% & +46\% \\
11--25 & +78\% & +2\% & +17\% \\
More than 25 & +68\% & \textless1\% & +7\% \\
\end{longtable}

\subsection{Reading the predicted
distribution}\label{sec-distribution-results}

\begin{figure}[H]

\centering{

\includegraphics[width=0.9\linewidth,height=\textheight,keepaspectratio]{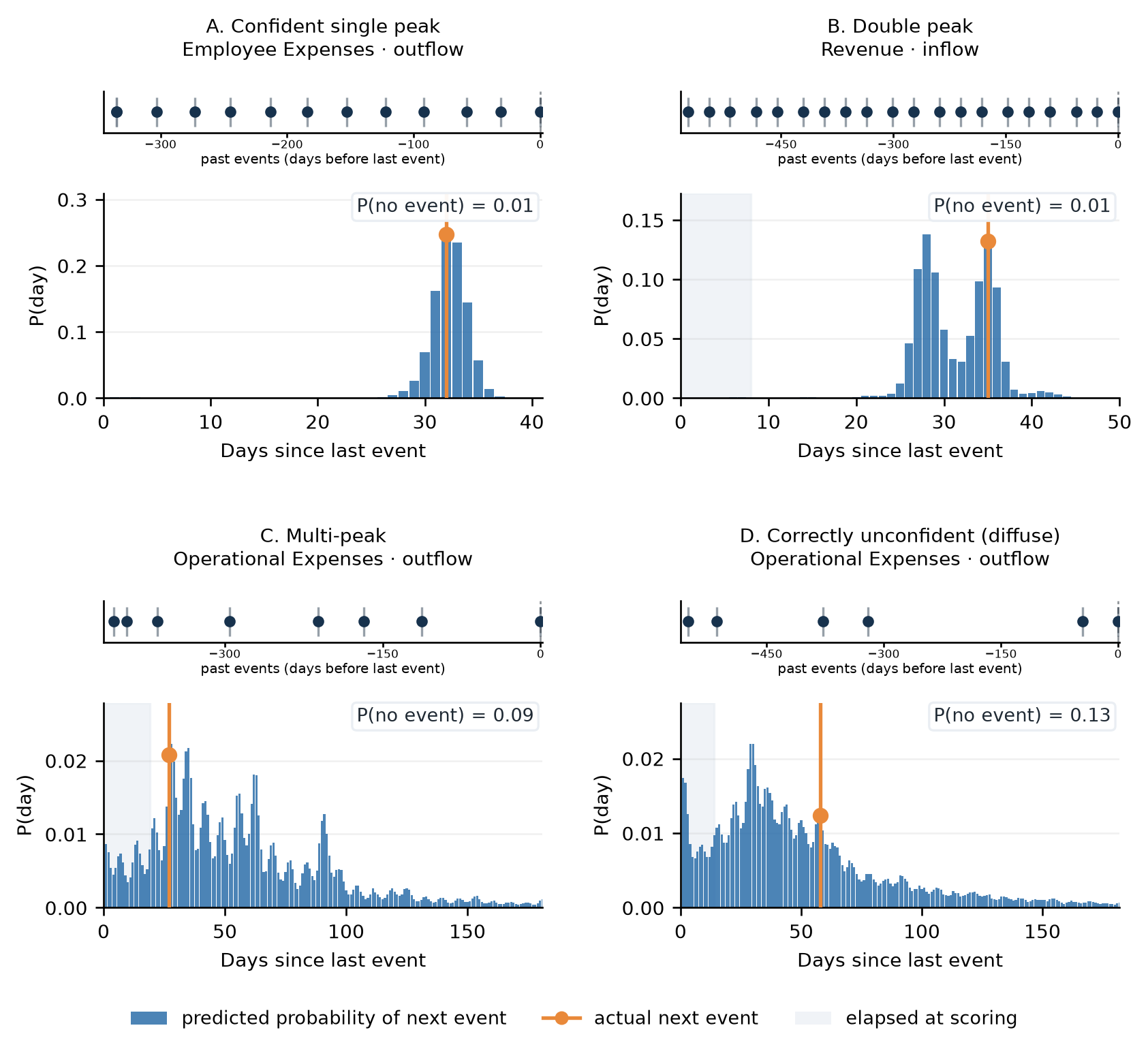}

}

\caption{\label{fig-distribution-gallery}Four shapes of the predicted
next-event distribution on validation streams: (A) a monthly payment
shifted past a weekend, (B) a second-Wednesday-of-the-month payment with
peaks at four and five weeks, (C) a midweek expense with irregular
spacing and peaks seven days apart, and (D) an irregular stream on which
the model is correctly unconfident. The strip above each panel shows the
stream's past events relative to its last event, bars show the predicted
probability of the next event on each day, gray shading marks days that
had already elapsed without an event at the reference date, and the
orange marker shows when the next event actually occurred.}

\end{figure}

Because CADENCE returns a distribution rather than a date, the shape of
each prediction describes how a stream recurs, not only when it pays
next. Figure~\ref{fig-distribution-gallery} shows four validation
streams with characteristic shapes, each reflecting calendar structure
that neither a fixed interval nor a single parametric curve can express.

\begin{itemize}
\tightlist
\item
  \textbf{(A) Confident single peak: a stable schedule.} An
  employee-expense outflow is paid on the 2nd of each month and moved to
  the next business day when the 2nd falls on a weekend; the 2nd of the
  target month is a Saturday, and the model's peak falls on the
  following Monday, when the payment arrived.
\item
  \textbf{(B) Double peak: a choice between schedules.} A revenue stream
  pays on the second Wednesday of every month, so the next payment
  arrives either four or five weeks after the previous one; the model
  splits its probability between 28 and 35 days, and the payment arrived
  at 35.
\item
  \textbf{(C) Multi-peak: a weekday anchor with irregular spacing.} An
  expense stream almost always posts on a Tuesday or Wednesday, at
  intervals from under two weeks to nearly four months, and the model
  places its peaks seven days apart, on successive occurrences of those
  weekdays.
\item
  \textbf{(D) Diffuse: no stable rhythm.} The history's gaps range from
  one to nine months, so the model is correctly unconfident: it spreads
  probability over several months and reserves more for no event at all,
  and the payment arrived inside that broad region but at no single
  predictable day.
\end{itemize}

Shape is a usable confidence signal on its own. Across validation
streams, predictions with a single sharp peak land within one day of
their most likely date nearly ten times as often as diffuse predictions,
with double- and multi-peak predictions in between.

\begin{figure}[H]

\centering{

\includegraphics[width=1\linewidth,height=\textheight,keepaspectratio]{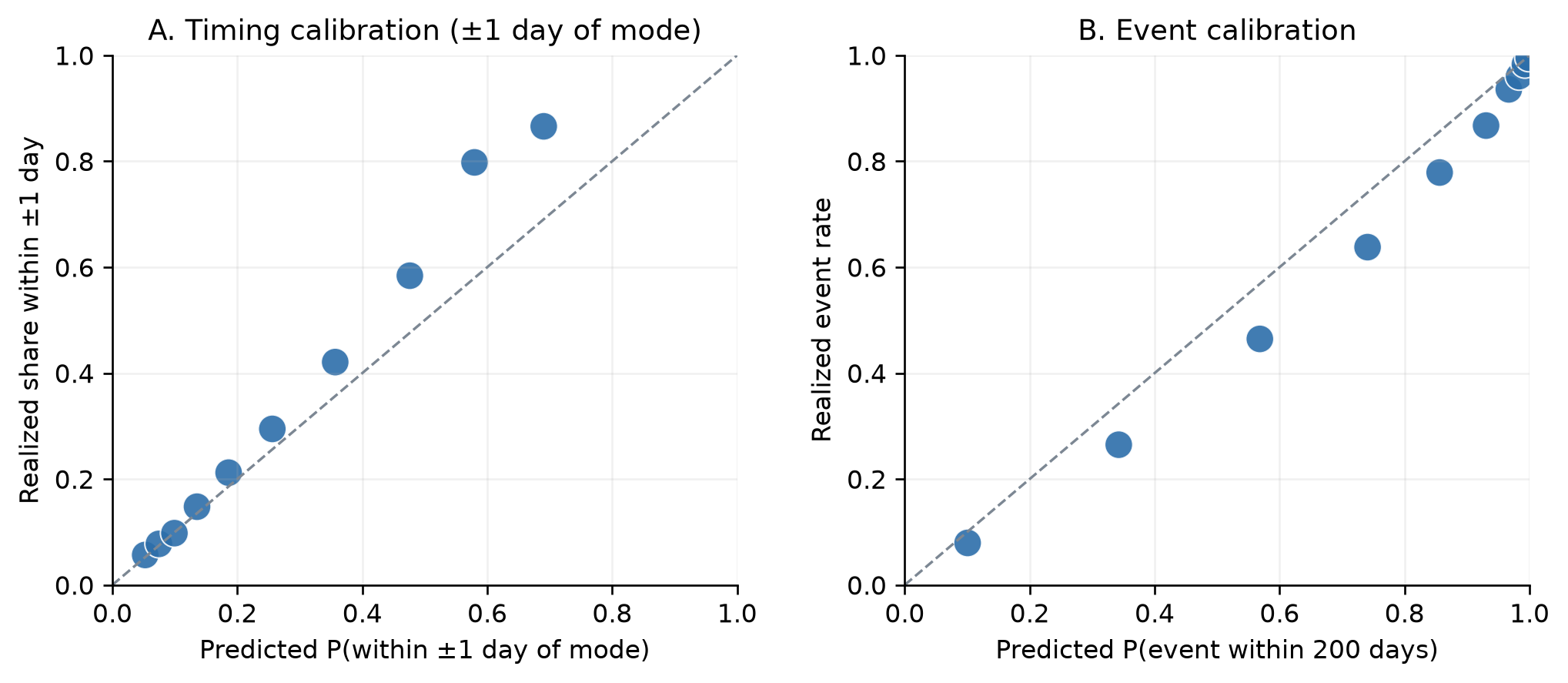}

}

\caption{\label{fig-calibration}Calibration of the predicted
distribution on validation streams. Left: predicted versus realized
probability that the next event lands within one day of the most likely
date. Right: predicted versus realized probability that a next event
occurs within the 200-day horizon. The diagonal marks perfect
calibration.}

\end{figure}

The probabilities themselves are also calibrated, in the sense that
stated confidence tracks realized frequency
(Figure~\ref{fig-calibration}). For timing, we bin predictions by the
probability the model assigns to the next event landing within one day
of its most likely date and compare it with how often that happened; for
occurrence, we compare the predicted probability that any next event
occurs within the horizon with how often one does. Both panels first
condition each distribution on the days that had already passed without
an event at the reference date, an update that a full distribution
permits by simple renormalization and a point forecast does not.
Conditioning matters: it reduces the occurrence calibration error about
fivefold, because a stream that has already been silent for weeks is
less likely to pay than its distribution at the last event implied.
After conditioning, the expected calibration error is about 0.07 for
timing and 0.05 for occurrence, and realized accuracy rises
monotonically with predicted confidence in both panels. FFT offers no
comparable property: its strength and regularity scores rank streams,
but they do not correspond to the probability of any observable outcome.

\subsection{Choosing the prediction horizon}\label{sec-horizon-results}

\begin{figure}[H]

\centering{

\includegraphics[width=1\linewidth,height=\textheight,keepaspectratio]{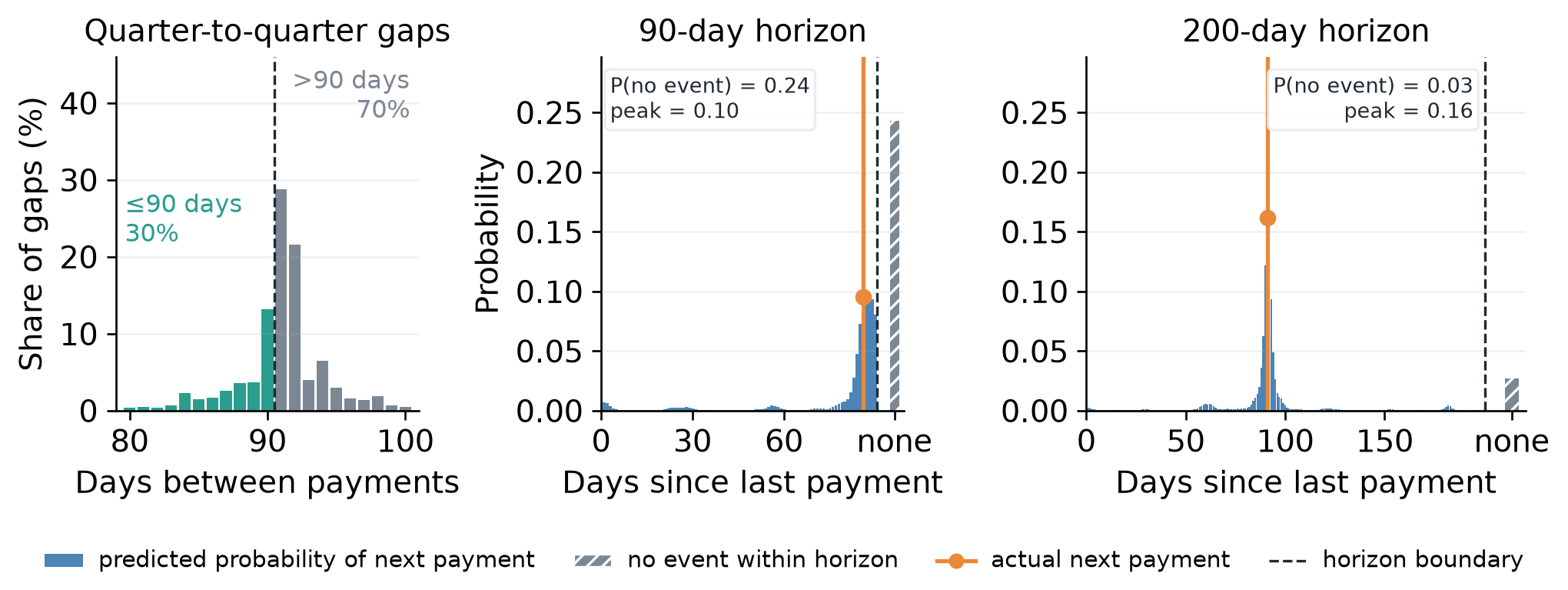}

}

\caption{\label{fig-horizon-boundary}Why the forecast horizon must clear
the quarter. Left: on-schedule gaps of 80 to 100 days between
consecutive payments in regular quarterly streams; because three
calendar months span 89 to 92 days, 70\% exceed 90 days. Middle: a
quarterly sales-tax stream scored by an earlier model with a 90-day
horizon, whose daily distribution peaks one day from the actual payment
while the pooled no-event outcome outranks every individual day. Right:
a quarterly sales-tax stream scored by CADENCE with a 200-day horizon,
which places its peak on the day of the actual payment, 91 days after
the last, and leaves little probability for no event.}

\end{figure}

The same distributions explain a design decision that shaped the whole
model: how far ahead it should look. An earlier version of the model
forecast 90 days ahead, and its failures on quarterly streams were
systematic rather than random (Figure~\ref{fig-horizon-boundary}). The
cause is calendrical: three calendar months span 89 to 92 days, so a
90-day boundary cuts through the middle of the quarterly cadence. Among
regular quarterly streams in the training data, whose median gap lies
between 84 and 98 days with little spread across at least four payments,
70\% of on-schedule gaps (80 to 100 days) exceed 90 days, and under a
90-day horizon every one of them becomes a no-event example.

The model learns this labeling faithfully. For regular quarterly streams
whose next payment arrived on schedule, 80 to 100 days after the last,
the 90-day model ranked no event above every individual day in all 46
validation cases, with an average no-event probability of 0.61. Its
day-level distribution nonetheless peaked within three days of the
realized date in two cases out of three. The model knew when the payment
would come; the no-event class, which pools every longer gap into a
single outcome, simply outweighed that knowledge.

Extending the horizon to 200 days moves quarterly and semiannual
schedules comfortably inside the window. On the corresponding validation
streams, CADENCE assigns an average no-event probability of 0.05, ranks
no event first in fewer than one case in five, and places its peak
within three days of the realized date in 71\% of cases. The right panel
of Figure~\ref{fig-horizon-boundary} shows a payment that arrived 91
days after the last, precisely on the model's peak, which a 90-day
horizon would have recorded as no event. With the longer horizon, the
no-event class means what underwriting needs it to mean, that a stream
has fallen silent for about six and a half months, rather than that a
quarterly payment landed a day or two past an arbitrary cutoff.

\subsection{Ablations}\label{sec-ablation-results}

\begin{figure}[H]

\centering{

\includegraphics[width=1\linewidth,height=\textheight,keepaspectratio]{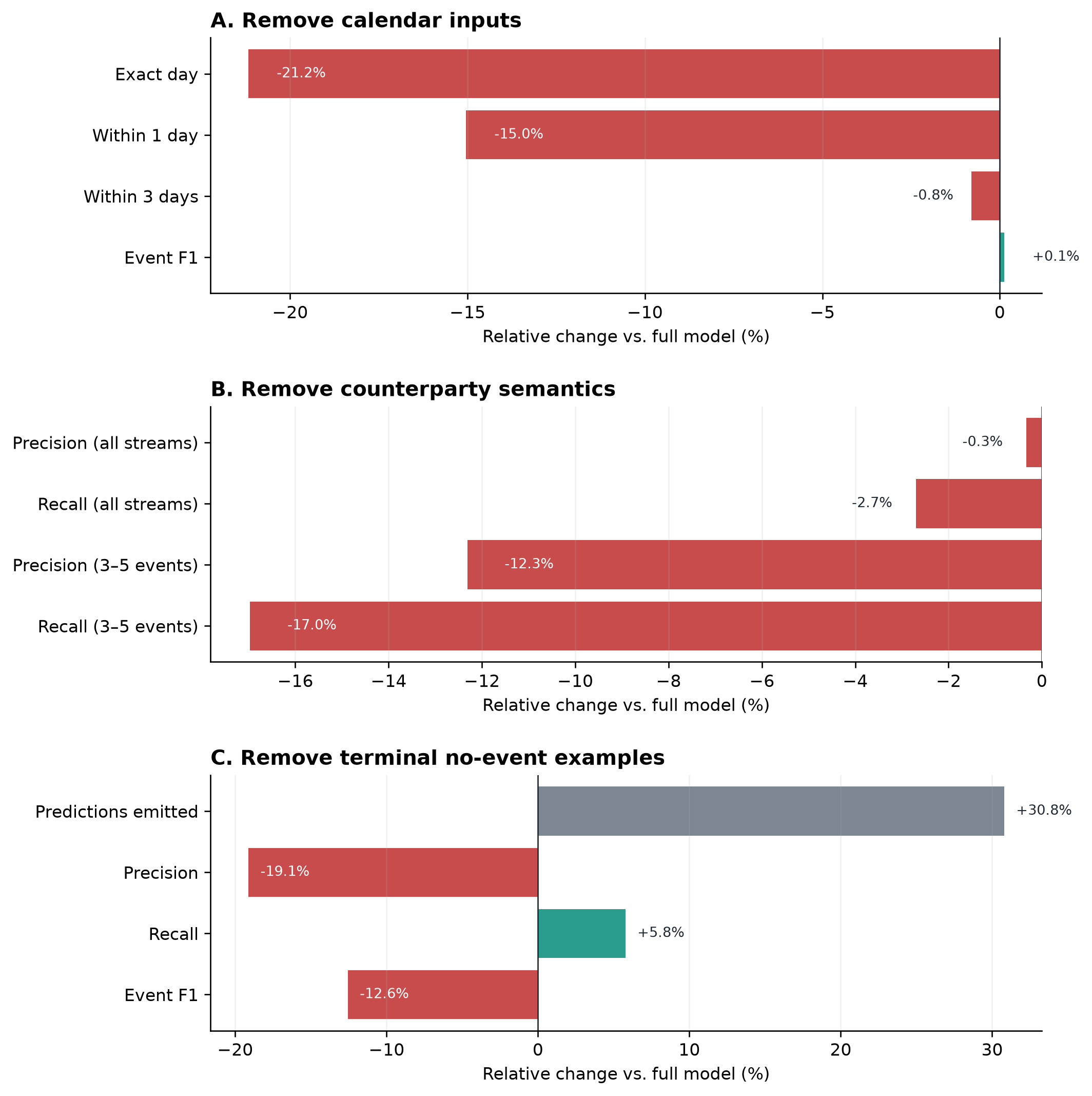}

}

\caption{\label{fig-ablation-results}Selected ablations relative to the
full CADENCE model. Calendar inputs drive exact-day timing, counterparty
semantics drive short-history recall, and no-event supervision drives
precision.}

\end{figure}

We ablate the three design choices whose removal produces a material and
interpretable change, removing one input or training mechanism at a time
and leaving the model otherwise identical
(Table~\ref{tbl-ablation-summary}, Figure~\ref{fig-ablation-results}).

\begin{longtable}[]{@{}
  >{\raggedright\arraybackslash}p{(\linewidth - 4\tabcolsep) * \real{0.4200}}
  >{\raggedright\arraybackslash}p{(\linewidth - 4\tabcolsep) * \real{0.4200}}
  >{\raggedleft\arraybackslash}p{(\linewidth - 4\tabcolsep) * \real{0.1600}}@{}}
\caption{Key effect of each ablation relative to the full CADENCE
model.}\label{tbl-ablation-summary}\tabularnewline
\toprule\noalign{}
\begin{minipage}[b]{\linewidth}\raggedright
Ablation
\end{minipage} & \begin{minipage}[b]{\linewidth}\raggedright
Key metric
\end{minipage} & \begin{minipage}[b]{\linewidth}\raggedleft
Relative change
\end{minipage} \\
\midrule\noalign{}
\endfirsthead
\toprule\noalign{}
\begin{minipage}[b]{\linewidth}\raggedright
Ablation
\end{minipage} & \begin{minipage}[b]{\linewidth}\raggedright
Key metric
\end{minipage} & \begin{minipage}[b]{\linewidth}\raggedleft
Relative change
\end{minipage} \\
\midrule\noalign{}
\endhead
\bottomrule\noalign{}
\endlastfoot
Remove calendar inputs & Exact-day timing accuracy & −21\% \\
Remove counterparty semantics & Recall on streams with 3--5 events &
−17\% \\
Remove terminal no-event examples & Next-event precision & −19\% \\
\end{longtable}

\textbf{Calendar inputs drive exact-day timing.} Removing the calendar
inputs reduces exact-day accuracy by about 21\% and one-day accuracy by
about 15\%, while three-day accuracy changes by less than 1\% and
next-event detection is essentially unchanged. Without calendar inputs,
the model still knows that a stream recurs and roughly when, but it can
no longer pick the specific day when that day is set by the calendar
rather than by the elapsed interval.

\textbf{Counterparty semantics drive short-history recall.} Removing the
frozen counterparty embedding barely changes aggregate precision, but
next-event recall falls by 2.7\%. The effect concentrates in streams
with three to five events, where next-event recall falls by about 17\%
and precision by about 12\%. With a long history, timing alone can
establish recurrence. With a short one, knowing that the counterparty is
a payroll processor, a landlord, or a lender supplies evidence that
spacing cannot, which is exactly the regime of newly recurring streams
that FFT misses.

\textbf{No-event supervision drives precision.} Removing the terminal
no-event examples leads the model to emit 31\% more predictions. Recall
rises by 6\%, but precision falls by 19\% and next-event F1 by 13\%.
Without examples of streams that end, the model learns that every
pattern continues, and the no-event class loses its meaning. Terminal
examples make abstention a learned outcome rather than a post-processing
rule.

\subsection{Downstream credit impact}\label{sec-downstream-results}

\begin{figure}[H]

\centering{

\includegraphics[width=0.78\linewidth,height=\textheight,keepaspectratio]{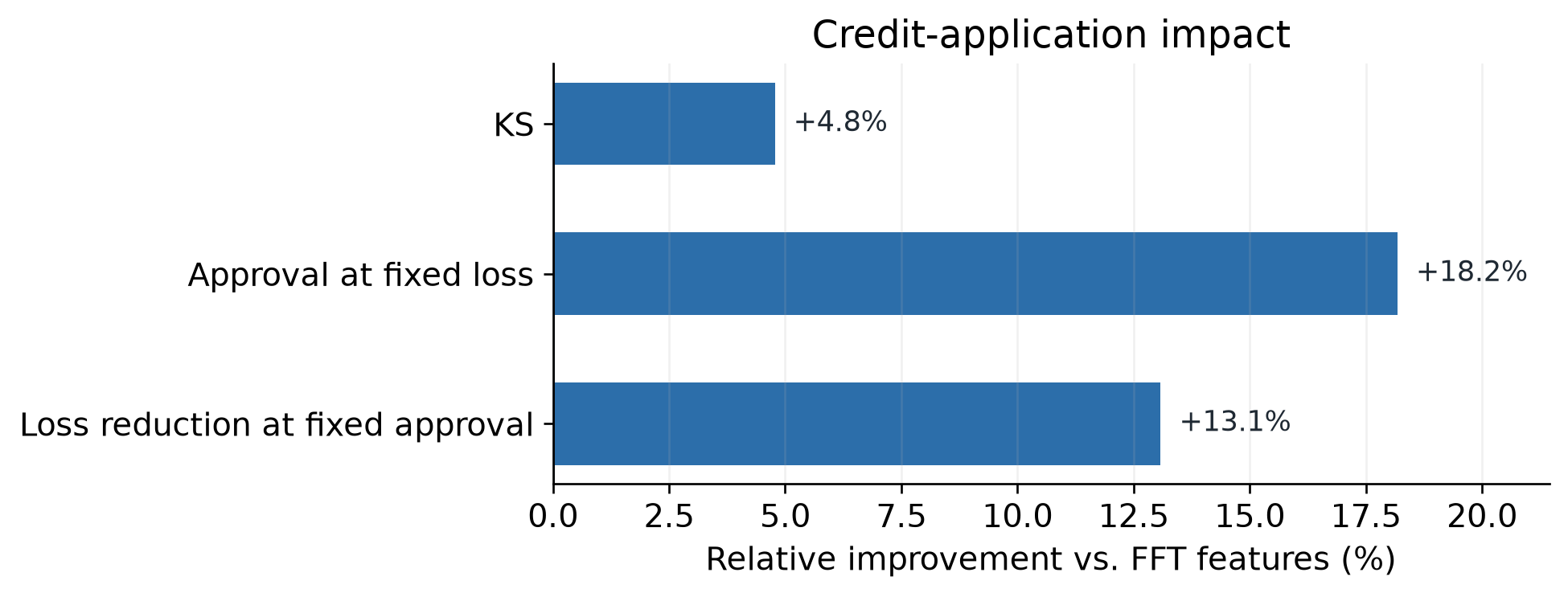}

}

\caption{\label{fig-downstream-results}Downstream credit-application
lift from CADENCE recurrence features relative to FFT recurrence
features.}

\end{figure}

Replacing FFT recurrence features with CADENCE features improves all
three decision metrics in the credit-application evaluation
(Figure~\ref{fig-downstream-results}). KS improves by 4.8\%, the
approval rate at a fixed loss rate rises by 18.2\%, and the loss rate at
a fixed approval rate falls by 13.1\%. In lending terms, the same loss
budget supports 18.2\% more approved applicants, or the same approval
volume carries 13.1\% fewer losses.

Recurrence features are one input family among many, so these gains show
that CADENCE's outputs carry information the credit model can use
alongside the rest of the underwriting data. The approval and loss
metrics measure that effect at the approval threshold, where lending
decisions are actually made, rather than only in overall ranking.

\section{Discussion}\label{sec-discussion}

\subsection{Priors substitute for scale}\label{sec-semantics-help}

The central lesson is that, at the data scale of business lending,
inherited priors do the work that scale does for foundation models.
PRAGMA's recurring-transaction task gained little from a hundredfold
increase in parameters or from adding a pre-trained text encoder, which
suggests that with 24 billion events a model can learn what it needs
about recurrence from the data itself (Ostroukhov et al. 2026). CADENCE
operates in the opposite regime: with most streams short and a long tail
of rarely seen counterparties, removing semantic priors costs 17\% of
recall on short histories, and short histories are exactly where
CADENCE's advantage over FFT is largest. The model does not need to
learn that a counterparty is a payroll processor, because the labeler
and the text encoder already know; it needs only to learn how payroll
behaves over time. That division of labor is how a 2.7-million-parameter
model trained on a single GPU can outperform a mature production method,
and it suggests that the approach suits business lending broadly, where
data is similarly limited.

\subsection{Why calendar awareness matters}\label{sec-calendar-timing}

Calendar awareness enters CADENCE in two places, its inputs and its
formulation, and the results speak to both. On the input side, three
separate results point to the same conclusion. Removing calendar inputs
costs exact-day accuracy but leaves detection unchanged, unlike the
other inputs we ablated (Section~\ref{sec-ablation-results}). Timing
improves over FFT at every history length, including long, regular
streams that FFT already detects reliably
(Table~\ref{tbl-history-length}). And the predicted distributions shift
a monthly payment past a weekend, split a second-Wednesday payment
between four and five weeks, and place peaks on successive occurrences
of the same weekday, rather than centering on a fixed interval
(Figure~\ref{fig-distribution-gallery}). Calendar inputs therefore solve
a timing problem, which is consistent with how business payments are
scheduled. Payroll, rent, loan installments, and many subscriptions are
anchored to calendar positions, such as every other Friday, the 1st and
15th, or the last business day, and they move around weekends and bank
holidays. Interval-based methods, including FFT, see these schedules as
fixed periods plus noise, which is enough to know that a stream recurs
but not to know the day. For applications that depend on the exact day,
such as projecting the lowest balance in the coming week or deciding
when a payment is late, that difference separates a usable forecast from
a rough one.

On the formulation side, the calendar dictates how far ahead the model
must look. Because three calendar months span 89 to 92 days, a horizon
that ends at 90 days assigns most quarterly payments to the no-event
class, however well the model knows their timing
(Section~\ref{sec-horizon-results}). The general rule is that a
categorical horizon should clear the longest cadence it is meant to
forecast by more than that cadence's calendar jitter, and that the
no-event class should be reserved for silences long enough to mean that
a stream has actually stopped.

\subsection{Interpretable by
construction}\label{sec-distribution-discussion}

Transaction foundation models produce embeddings whose meaning is
defined only by the downstream models trained on them; CADENCE produces
a distribution whose every value is the probability of an observable
event. Because that distribution is nonparametric, it also records how a
stream actually recurs, whether skewed, long-tailed, or multi-peaked,
rather than forcing it into one period or one bell-shaped curve. In many
domains, the predictive distribution is a useful byproduct of the mean
prediction, but for recurrence, it is perhaps more useful than the
prediction itself. A single forward pass tells an underwriter whether a
stream continues, through the event probability, when it pays next,
through the peak, and how far to trust that answer, through the peak's
prominence and the entropy of the whole distribution. The forecast also
stays current as time passes without new data, since conditioning on the
days already elapsed is a simple renormalization that keeps occurrence
probabilities calibrated as a stream falls silent
(Section~\ref{sec-distribution-results}). The horizon study adds a
subtler point: the 90-day model knew when quarterly payments would
arrive, but a comparison between individual days and the pooled no-event
class threw that knowledge away (Section~\ref{sec-horizon-results}).
Decisions should therefore read the shape of the distribution rather
than only its most likely outcome, which is what the serving gate does
when it combines event probability with measures of peak sharpness and
spread, so that a stream with a sharp peak can be accepted at a lower
event probability than a diffuse one. These properties, more than any
single accuracy metric, are what make next-event forecasting a natural
successor to periodicity scores in underwriting.

\subsection{Applications to underwriting and risk
management}\label{sec-prediction-to-underwriting}

Better recurrence forecasts benefit credit decisions in several ways:

\begin{itemize}
\tightlist
\item
  \textbf{Income estimation.} Higher recall on new and short streams
  means recurring revenue is recognized sooner, which matters most for
  young businesses and recently connected accounts, whose histories are
  short by definition. Learned no-event predictions also keep ended
  streams from inflating recurring revenue, the error that most directly
  overstates a credit line. Because income estimates feed credit-line
  assignment and policy rules as well as risk models, these gains reach
  decisions beyond any single score.
\item
  \textbf{Cash-flow forecasting.} Stream-level distributions compose
  into a customer-level forecast: weighting each stream's day
  probabilities by its predicted amount and summing across all of a
  customer's streams yields an expected daily schedule of future inflows
  and outflows for the business as a whole. That schedule is the raw
  material for projected balances and debt-service coverage, and sharper
  timing improves exactly these estimates, which depend on when money
  arrives relative to when it leaves.
\item
  \textbf{Credit risk modeling.} CADENCE features improve downstream
  credit models (Section~\ref{sec-downstream-results}). Those models can
  consume not only more accurate revenue features but also new
  dimensions of recurrence, such as timing confidence and termination
  risk, that frequency-domain scores cannot provide.
\item
  \textbf{Anomaly detection.} The same outputs support portfolio
  monitoring after origination. A confidently expected payroll run or
  revenue deposit that has not arrived after a grace period, or a new
  financing debit that starts to recur, is an early-warning signal that
  no periodicity score provides. Because the model's confidence tracks
  realized accuracy, a monitoring threshold can be stated as the
  probability that a payment was missed, and the grace period can follow
  the width of the predicted peak, rather than either being a hand-tuned
  rule.
\end{itemize}

\section{Limitations and Future Work}\label{sec-limitations}

\textbf{Scope of evidence.} Direct metrics are measured on customers
disjoint from training at historical reference dates, and downstream
metrics within the observed underwriting population. Neither establishes
causal business impact or prescribes a credit policy. Prospective
monitoring on newly originated applications is the definitive test of
sustained lift.

\textbf{Drift.} Counterparty naming, payment rails, and customer mix
change over time, and the model inherits any drift in the upstream
labeler. Periodic retraining and monitoring of prediction volume, timing
error, and feature distributions are therefore necessary.

\textbf{Amounts and cross-stream context.} The amount head produces a
point estimate, and each prediction conditions on a single stream.
Distributional amount predictions would support cash-flow forecasts with
uncertainty bands, and attention across a customer's streams could
capture dependencies such as payroll that follows revenue.

\textbf{Responsible use.} The model supplies features to underwriting
and does not make credit decisions. Its effect on approvals and losses
should be assessed, monitored, and governed as part of the complete
decision process, including the fairness and explainability obligations
that apply to that process.

\section{Conclusion}\label{sec-conclusion}

Identifying recurring cash flows is essential to business underwriting
but difficult in practice, because so much business income arrives as
ambiguous transfers on short, calendar-anchored schedules. CADENCE
replaces frequency-domain recurrence estimation with a compact causal
Transformer that treats each transaction as a token, inherits semantics
from an LLM labeler and a frozen text encoder, and predicts a full,
nonparametric distribution over the next event. It improves one-day
timing accuracy by 93.3\% and next-event F1 by 6.7\%, and it recognizes
short streams that FFT largely misses. Its distributions follow the
skewed, multi-peaked structure of real cash-flow timing and carry
calibrated confidence. Calendar inputs drive exact-day timing,
counterparty semantics drive short-history recall, and no-event
supervision drives precision. In an end-to-end credit evaluation,
CADENCE features raise KS by 4.8\%, approve 18.2\% more applicants at
the same loss rate, and lower the loss rate by 13.1\% at the same
approval rate. When data is limited and outputs must be interpretable
and compatible with existing credit systems, a small, specialized model
with strong priors and a distributional output is a practical
alternative to scale-first transaction foundation models.

\section*{References}\label{references}
\addcontentsline{toc}{section}{References}

\protect\phantomsection\label{refs}

\end{document}